\documentclass[letterpaper, 10 pt, conference]{ieeeconf}  %

\IEEEoverridecommandlockouts                              %

\usepackage[tracking=true]{microtype}
\usepackage{amsmath} %
\usepackage{amssymb}  %
\usepackage{graphicx}
\usepackage{epsfig} %
\usepackage{mathptmx} %
\usepackage{times} %
\usepackage{amsmath} %
\usepackage{amssymb}  %
\usepackage[frozencache,cachedir=minted]{minted}
\usepackage{minted}
\setminted{fontsize=\footnotesize}
\usepackage{url}
\usepackage[dvipsnames]{xcolor}
\usepackage[font={footnotesize}]{caption}
\usepackage[caption=false, font=footnotesize]{subfig}
\usepackage{booktabs} %
\usepackage{tikz}
\usetikzlibrary{positioning,shapes.geometric,arrows,fit,shapes.symbols,svg.path,tikzmark,calc}
\usepackage{booktabs}
\usepackage{textcomp}
\usepackage[normalem]{ulem} %
\let\labelindent\relax
\usepackage[inline]{enumitem} %
\usepackage{multirow}
\usepackage{hyperref}

\pgfdeclarelayer{background}
\pgfdeclarelayer{foreground}
\pgfsetlayers{background,main,foreground}

\newcommand{\FogROStwo}{FogROS2}

\newcommand{\ROS}[1]{ROS\,#1}

\newcommand{\remark}[3]{\textcolor[hsb]{#1,1.0,0.8}{[#3 -#2]}}

\newcommand{\joey}[1]{\remark{0.8}{JG}{#1}}

\renewcommand{\joey}[1]{}

\title{\LARGE \bf
FogROS2: An Adaptive Platform \\ for Cloud and Fog Robotics Using ROS 2
}

\author{Jeffrey Ichnowski$^{*1A,3}$, Kaiyuan Chen$^{*1}$, Karthik Dharmarajan$^{1A}$, Simeon Adebola$^{1A}$, 
  Michael Danielczuk$^{1A}$, \\ 
  Víctor Mayoral-Vilches$^{4,5}$, Nikhil Jha$^{1A}$, 
  Hugo Zhan$^{1A}$, Edith LLontop$^{1A}$, Derek Xu$^{1A}$, Camilo Buscaron, \\ %
  John Kubiatowicz$^{1}$, Ion Stoica$^{1}$, Joseph Gonzalez$^{1}$, and Ken Goldberg$^{1,2,A}$%
\thanks{$^{*}$Equal Contribution}
\thanks{$^{1}$Department of Electrical Engineering and Computer Science}%
\thanks{$^{A}$The AUTOLab at UC Berkeley (\href{http://automation.berkeley.edu}{automation.berkeley.edu}).}
\thanks{$^{2}$Department of Industrial Engineering and Operations Research}%
\thanks{$^{1,2}$University of California, Berkeley, CA, USA }%
\thanks{$^{3}$Robotics Institute, Carnegie Mellon University}%
\thanks{$^{4}$Acceleration Robotics, Ecuador 3, 1 I, Vitoria, Álava, Spain, }%
\thanks{$^{5}$System Security Group, Universit\"at Klagenfurt, Universitätsstr. 65-67 9020 Klagenfurt, Austria}%
\thanks{{\tt\footnotesize jeffi@cmu.edu, kych%
@berkeley.edu}}%
}

\pdfmapfile{=cid-base.map}

\begin{document}

\maketitle
\thispagestyle{empty}
\pagestyle{empty}

\begin{abstract}
Mobility, power, and price points often dictate that robots do not have sufficient computing power on board to run contemporary robot algorithms at desired rates.  Cloud computing providers such as AWS, GCP, and Azure offer immense computing power and increasingly low latency on demand, but tapping into that power from a robot is non-trivial.  We present FogROS2, an open-source platform to facilitate cloud and fog robotics that is included in the Robot Operating System 2 (ROS 2) distribution.  FogROS2 is distinct from its predecessor FogROS1 in 9 ways, including lower latency, overhead, and startup times; improved usability, and additional automation, such as region and computer type selection.  Additionally, FogROS2 gains performance, timing, and additional improvements associated with ROS 2. In common robot applications, FogROS2 reduces SLAM latency by 50\,\%, reduces grasp planning time from 14\,s to 1.2\,s, and speeds up motion planning 45x.  When compared to FogROS1, FogROS2 reduces network utilization by up to 3.8x, improves startup time by 63\,\%, and network round-trip latency by 97\,\% for images using video compression. The source code, examples, and documentation for FogROS2 are available at \url{https://github.com/BerkeleyAutomation/FogROS2}, and is available through the official ROS 2 repository at \url{https://index.ros.org/p/fogros2/}.

\end{abstract}

\section{Introduction}
\label{sec:introduction}

\begin{figure}[t]
    \centering
      \input{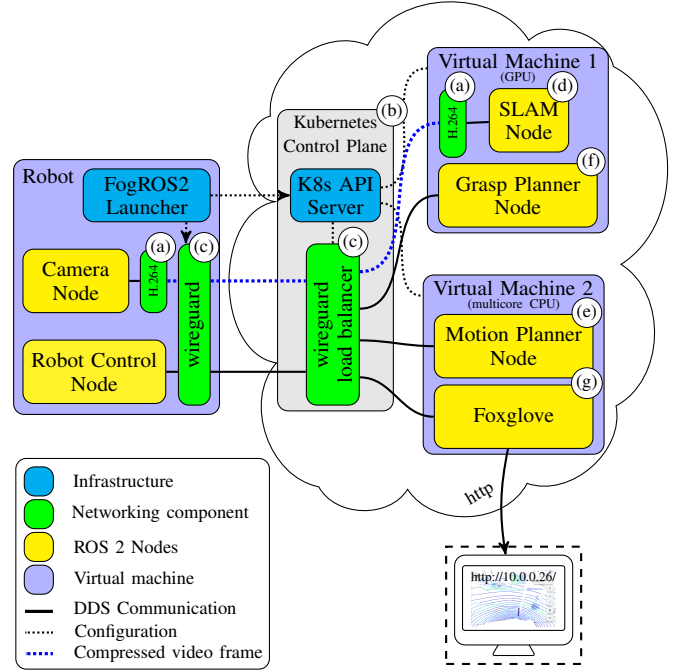}%
    \caption {%
    \FogROStwo{} addresses several deficiencies found in FogROS1, including latency, usability, and automation.  For example, 
    transmitting images from a camera to the cloud can require a lot of bandwidth, thus increasing latency. (a) \FogROStwo{} transparently compresses video and image streams using the popular H.264 compression standard.
    \FogROStwo{} further improves upon FogROS1 by speeding up launch times, improving scalability, and increasing compatibility by (b) migrating to Kubernetes (K8s), (c) switching to UDP over Wireguard, and updating to the ROS 2 ecosystem. 
    In an example application, (d) \FogROStwo{} moves a simultaneous localization and mapping (SLAM), (e) motion planning, (f) and grasp planning nodes to the cloud, taking advantage of the GPU and high CPU core counts available there to speed up processing. (g) By integrating Foxglove, users are also able to monitor robots from a browser any where in the world.\vspace{-8pt}
    }
    \label{fig:vpc_overview_2}
\end{figure}

It is difficult for the onboard computing resources of robots to keep up with advances in robot algorithms and new computing hardware.  %
Cloud computing offers on-demand access to immense computing resources and new and power-hungry computing platforms, such as GPUs, TPUs, and FPGAs. %
Prior work~\cite{ichnowski2017economic} showed that using the cloud for intensive computing in robotics can be practical and cost-effective. %
However, gaining access to evolving cloud computing resources requires expertise with many new and emerging software packages, and experience handling data security and privacy. %
In prior work~\cite{chen2021fogros}, we introduced FogROS (henceforth \emph{FogROS1}), a framework that extends the Robot Operating System (ROS) (henceforth \emph{\ROS1}) to enable quick access to the cloud.

However, FogROS1 has limitations in latency, usability, and automation. 
In this paper, we introduce \FogROStwo{} to reduce latency, improve usabillity, and automate additional components of launching robot code in the cloud, while extending to additional potential robot applications.  Furthermore, we rewrote \FogROStwo{} from scratch to fully integrate with \ROS2 to benefit from improvements in networking, launch configurability, and its command line interface; and we added integration points to Foxglove~\cite{foxglove} to enable \emph{remote} monitoring from anywhere in the world.

Latency, the time between when an event occurs and when the robot reacts to the event, is a critical factor in many applications.  FogROS1 and \FogROStwo{} demonstrate that using the cloud can reduce latency of complex computations.  (It is important to note that this is not a universal statement---some computation, such as feedback control loops, time-bounded, or safety-critical computations are not always suitable for the cloud due to unpredictable network round-trip times.)
FogROS1 suffered from long cloud computer startup times (around 4 minutes) and high round-trip network times, particularly for images (e.g., 5 seconds per image).  FogROS2 lowers these latencies by using application-specific cloud-computer images, using a Kubernetes backend to avoid overhead associated with creating new cloud computers, switching from TCP to UDP secured networking, and adding transparent H.264 video compression for image topics.

\FogROStwo{} includes several usability and automation improvements over FogROS1 to facilitate adoption, including: (1) a command-line interface (CLI) to interact with \FogROStwo{} cloud computers, (2) integration with Foxglove to enable remote monitoring, (3) a new launch process to allow for automating cloud-computer specification and region, and (4) creation of custom cloud-computer images.

Many of these improvements are facilitated by re-writing \FogROStwo{} for \ROS2.
\ROS2~\cite{macenski2022robot}, a rapidly growing replacement for \ROS1~\cite{quigley2009ros}, is a standard for developing robot applications.
FogROS1 and \FogROStwo{} enable moving computationally intensive parts (or \emph{nodes}) of a computational graph to the cloud and securing communication channels for messages, all with a few small changes to the launch script and without changing a line of the robot code.
By migrating to ROS\,2, the \FogROStwo{} launch system gains additional capabilities, such as: detecting the cloud server region nearest to the robot and automatically selecting computers and images based computational requirements. %
\FogROStwo{} is now part of the official ROS 2 ecosystem, and installable with standard Ubuntu commands (\texttt{\footnotesize apt install ros-humble-fogros2}).

In 3 example applications, visual SLAM, grasp planning, and motion planning, we evaluate the ability of \FogROStwo{} to reduce total computation times.  
We find that using
cloud computers via \FogROStwo{} speeds up the
computation and
can reduce compute times by 2x to 45x. %
Comparing to
FogROS1, \FogROStwo{} improves startup times by 63\,\%, %
and network latency by 97\,\% for images.

\FogROStwo{} contributes 9 improvements over FogROS1~\cite{chen2021fogros}:
\begin{enumerate*}[label=(\arabic*)]
\item \FogROStwo{} extends the ROS\,2 launch system introducing additional syntax in launch files that allow roboticists to specify at launch time which components of their architecture will be deployed to the cloud, and which ones on the robot. While FogROS1 existed outside the official ROS ecosystem, \FogROStwo{} directly integrates with it. %
\item \FogROStwo{} provides launch script logic that allows robots to automate selection of cloud-computing resources, such as nearest region, computer image, and computer type.
\item \FogROStwo{} adds support for streaming video compression between robot and cloud nodes---significantly improving the performance of image processing in the cloud, and potentially enabling new applications. \joey{Engineering?  Why is this not in FogROS1?}
\item The architecture of \FogROStwo{} is extensible, making it easy to plug in support for additional cloud computing providers, Data Distribution Service (DDS) providers (Sec.~\ref{sec:dds}), and message compression. 
\item \FogROStwo{} integrates with \ROS2 tooling and provides \ROS2 command-line interfaces to query and control cloud-robotics deployments. %
\item \FogROStwo{} interfaces with the new Foxglove web-based robot visualization software~\cite{foxglove} to allow remote (anywhere-in-the-world) monitoring of \FogROStwo{} applications. %
\item \FogROStwo{} supports a new backend based on Kubernetes that allows for faster warm starts and broader cloud-service provider support. %
\item \FogROStwo{} automates the building of cloud-based virtual machine images for faster startup time.
\item \FogROStwo{} is part of the \ROS{2} ecosystem and is accessible with the standard \texttt{apt install} command.
\end{enumerate*}

\section{Related Work}
\label{sec:related_work}

Robots have limited onboard computing capabilities and as the computing demands of robotics %
grow, the cloud has become an increasingly necessary source of computing power.  
Kehoe et al.~\cite{kehoe2015survey, kehoe2013cloud} surveys the landscape of cloud robotics, including capabilities, potential applications, and challenges.

Cloud-robotics platforms facilitate offloading computation and data to the cloud.  A notable example is RoboEarth~\cite{waibel2011roboearth}, which shared information between robot and cloud.  The main use case was to use the cloud to share databases between robots, but it did not leverage the cloud for offloading computing.  Rapyuta~\cite{mohanarajah2014rapyuta} emerged from RobotEarth to become a platform for centralized management of robot fleets.  In Rapyuta, robot nodes or Docker images are built on the cloud and pushed to the registered robots.  A similar approach is taken by AWS Greengrass~\cite{greengrass}.
Using proprietary interfaces, Rapyuta and Greengrass allow building and deploying an entire pipeline for robotics applications~\cite{lam2014path, mohanarajah2015cloud, mouradian2018robots, rosa2017exportation} from a centralized cloud interface. %
In contrast, \FogROStwo{} approaches cloud deployment from the opposite perspective---instead of pushing applications from the cloud to a robot, \FogROStwo{} pushes robot nodes from robot to the cloud. It uses an interface familiar to ROS\,2 developers, allowing developers and researchers to access cloud resources without learning or conforming to an additional framework. 

Researchers have explored using the cloud for grasp planning (e.g., Kehoe et al.~\cite{kehoe2013cloud}, Tian et al.~\cite{tian2017cloud}, and Li et al.~\cite{li2018dex}), parallelized Monte-Carlo grasp perturbation sampling~\cite{kehoe2012estimating,kehoe2012toward,kehoe2014cloud}, motion planning services (e.g., Lam et al.~\cite{lam2014path}), and splitting motion plan computation between robot and cloud (e.g., Bekris et al.~\cite{bekris2015cloud} and Ichnowski et al.~\cite{ichnowski2016cloud}).  Researchers also have explored using new cloud computing paradigms as they emerge, such as serverless computing~\cite{ichnowski2020fog,anand2021serverless}, in which algorithms run (and are charged) for short bursts of intensive computing; while others have explored using the cloud to gain access to hardware accelerators such as FPGAs~\cite{murray2016robot}.
Others have explored some of these challenges, such as preserving privacy~\cite{mahler2016privacy} and sharing models between robots~\cite{tanwani2020rilaas}.
In many of these examples, using the cloud requires a custom one-off implementation or interfacing with a proprietary library.  \FogROStwo{} and ROS\,2 reduces this complexity. %

For a robot to gain access to cloud resources, it must provision a cloud computer and establish a network connection to it.  As robots operate in the physical world, the connection to the cloud must be secured. 
However, setting this up is an involved process, in some cases requiring 12 steps for configuration and 37 steps for verification~\cite{hajjaj2017establishing}.
Hajjaj et al.~\cite{hajjaj2017establishing} explored using SSH tunnelling for communication with ROS nodes running in the cloud.  However, SSH tunnels do not support UDP which is needed when using ROS\,2 Data Distribution Service (DDS) over UDP (while some DDS implementations support TCP, using TCP can introduce performance issues, and add unnecessary overhead for local communication).
Crick et al. proposed rosbridge~\cite{crick2012rosbridge}, Pereira et al.~\cite{pereira2019rosremote} proposed ROS Remote, and Xu et al. proposed MSA~\cite{xu2020cloud} as alternate ROS communication stacks with varying degrees of security and modifications required for their use in ROS applications.
Wan et al.~\cite{wan2016cloud} and Saha et al.~\cite{saha2018comprehensive} propose unifying robot-cloud communication.
Lim et al. proposed using VPNs~\cite{lim2019cloud}, and \FogROStwo{} builds on this approach.
\FogROStwo{} allows ROS\,2 applications to easily use the cloud without code modification, and with secured communication.
\section{Background on ROS}
\label{sec:background}
\label{sec:dds}

ROS\,2~\cite{macenski2022robot}, the successor to Robot Operating System (ROS\,1), includes many substantial improvements.
A core improvement in ROS\,2 is its change from a proprietary publication/subscription (pub/sub) system to the industry-standard middleware Data Distribution Service (DDS)~\cite{dds2015}.  DDS addresses robotics concerns %
such as providing real-time, high-performance, interoperable, and reliable communication~\cite{dds2015}.  As DDS is a specification, there are several implementations, and ROS 2 is agnostic to DDS implementation. %

 In ROS 2, computational units are abstracted into \emph{nodes} that communicate with each other via a pub/sub system.  Nodes subscribe to named \emph{topics} and receive \emph{messages} (data) as other nodes \emph{publish} them.  %
 In an example application (Fig.~\ref{fig:vpc_overview_2}), a camera node publishes images, a Simultaneous Location And Mapping (SLAM) node processes the images and publishes a location and map, a Motion Planner node receives the map and then computes and publishes a collision-free path, and a path following node drives the wheels to reach a target. %

When orchestrating a robot application, often multiple nodes must be launched simultaneously.  The ROS 2 launch system facilitates this by providing the ability to specify all required nodes, topic mappings, and relations between nodes in a single python script file.  Launching the robot application is then a matter of running the command:
\[
\text{\texttt{\footnotesize ros2 launch <package> <script>}}.
\] 
Listing~\ref{lst:launch}, without the circled \FogROStwo{} extensions, shows an example launch script that launches two nodes, a ``grasp\_motion'' and a ``grasp\_planner'' simultaneously.
\section{Approach}
\label{sec:approach}

\begin{listing}[t]
\inputminted[xleftmargin=20pt,linenos,fontsize=\scriptsize,escapeinside=||]{python}{listing/fogros2_launch_example.py}%
\begin{tikzpicture}[remember picture, overlay]
  \node [fit=(pic cs:machinestart)(pic cs:machineend),inner xsep=2pt, inner ysep=6pt, xshift=-1pt, yshift=2pt, rounded corners,draw=black] {};
  \node [fit=(pic cs:attrstart)(pic cs:attrend),inner xsep=2pt, inner ysep=6pt, xshift=-1pt, yshift=2pt, rounded corners,draw=black] {};
\end{tikzpicture}%
\vspace{-12pt}
\caption{\textbf{\FogROStwo{} Launch Script Example.}  This example launches two nodes.  Unlike \ROS1 and FogROS1, which used an XML launch file, \FogROStwo{}'s launch files are python scripts.
In this example, the \FogROStwo{} launch extensions are circled.  The first extension defines a machine on which to launch nodes.  The second tells \FogROStwo{} to launch the grasp planner node on that machine.}
\label{lst:launch}
\end{listing}

\begin{figure}[t]
    \centering
    \begin{tikzpicture}[node distance=6pt,>=stealth',font=\footnotesize,
   rosblock/.style={draw, rectangle, rounded corners, align=center, inner sep=3pt},
   roslabel/.style={inner sep=0pt, xshift=3.2pt},
   rosseq/.style={circle, inner sep=1pt, fill=white, draw=black, xshift=-2pt, yshift=-2pt, font=\scriptsize}]

\newcommand{\sep}{3pt}

\node[rosblock, text width=102pt, align=left, inner sep=4pt, minimum height=90pt, xshift=-1pt, fill=blue!30] (robot) {};
\node[roslabel, below=3.2pt of robot.north west, anchor=north west] { Robot };

\node[rosblock, text width=122pt, align=left, inner sep=4pt, minimum height=90pt, fill=blue!30, right=3.39 in of robot.north west, anchor=north east] (cloud) {};
\node[roslabel, below=3.2pt of cloud.north west, anchor=north west] { Kubernetes on Cloud Provider };

\draw let \p1 = (cloud.north west), \p2 = (cloud.south east) in
 node[rosblock, below=12pt of cloud.north west, anchor=north west, text width=(\x2 - \x1 -28), align=left, xshift=18pt, inner sep=3pt, fill=blue!20!cyan, minimum height=(\y1 - \y2 - 12pt - 3.2pt)] (vm) {};
\node [roslabel, below=3.2pt of vm.north west, anchor=north west] { Container };
 
\draw let \p1 = (vm.north west), \p2 = (vm.south east) in
 node[rosblock, text width=(\y1 - \y2 + 3pt - 3*3.2pt), above=3pt of vm.south west, anchor=north west, xshift=-14.2pt, yshift= -3pt,  rotate=90, fill=green] (vpn) {VPN };

\draw %
 node[rosblock, right=3.2pt of vpn.south east, anchor=north east, text width=30, fill=green, rotate=90, xshift=-12pt, yshift=-3pt] (dds) { DDS };
\node [rosblock, right=3.pt of dds.south east, anchor=north west, text width=80pt, minimum height=37pt, fill=green] (wksp) { }; 
\node [roslabel, below=3.2pt of wksp.north west, anchor=north west] { Workspace };

\draw let \p1 = (wksp.north west), \p2 = (wksp.south east) in node [rosblock, below=12pt of wksp.north west, anchor=north west, fill=yellow, xshift=3.2pt, minimum height=(\y1 - \y2 - 12pt - 0.2pt - 3.2pt), text width = 32.5pt] (node1) { \scriptsize Node \\[-2pt] C };
\draw let \p1 = (wksp.north west), \p2 = (wksp.south east) in node [rosblock, below=12pt of wksp.north east, anchor=north east, fill=yellow, xshift=-3.2pt, minimum height=(\y1 - \y2 - 12pt - 0.2pt - 3.2pt), text width = 32.5pt] (node2) { \scriptsize Node\\[-2pt] D };
 
 \node [rosblock, right=3.pt of dds.south east, anchor=north west, text width=95.5pt, minimum height=20pt, fill=yellow, yshift=-40pt, xshift=-15pt] (ros dep) { ROS + dependencies};

\draw let \p1 = (vm.north west), \p2 = (vm.south east) in
 node[rosblock, text width=(\y1 - \y2 + 3pt - 3*3.2pt), above=3pt of vm.south west, anchor=north west, xshift=-38pt, yshift= -3pt,  rotate=90, fill=green] (vpn) {VPN };
\draw %
 node[rosblock, right=4pt of vpn.south east, anchor=north east, text width=32, fill=green, rotate=90, xshift=-11.5pt, yshift=30pt] (dds2) { DDS };
 \node [rosblock, right=-14.pt of dds2.south west, anchor=north east, text width=68pt, minimum height=37pt, fill=blue!20!cyan, yshift =38pt] (wksp2) { }; 
 \node [roslabel, below=3.2pt of wksp2.north west, anchor=north west] { Workspace };
 
 \node [rosblock, right=3.pt of dds2.south east, anchor=north west, text width=82pt, minimum height=22pt, fill=yellow, yshift=-41pt, xshift=-91pt] (ros dep2) { ROS + dependencies};

\draw let \p1 = (wksp2.north west), \p2 = (wksp2.south east) in node [rosblock, below=12pt of wksp2.north west, anchor=north west, fill=yellow, xshift=3.2pt, minimum height=(\y1 - \y2 - 12pt - 0.2pt - 3.2pt)] (robot node1) { \scriptsize Node \\[-2pt] A };

\draw let \p1 = (wksp2.north west), \p2 = (wksp2.south east) in node [rosblock, below=12pt of wksp2.north west, anchor=north west, fill=yellow, xshift=(3.2pt + 16pt), minimum height=(\y1 - \y2 - 12pt - 0.2pt - 3.2pt)] (robot node2) { \scriptsize Node \\[-2pt] B };
 
\draw let \p1 = (wksp2.north west), \p2 = (wksp2.south east) in node [rosblock, below=12pt of wksp2.north west, anchor=north west, fill=yellow!80, xshift=(3.2pt + 32pt), minimum height=(\y1 - \y2 - 12pt - 0.2pt - 3.2pt)] { \scriptsize \color{gray} Node \\[-2pt] \color{gray} C };
\draw let \p1 = (wksp2.north west), \p2 = (wksp2.south east) in node [rosblock, below=12pt of wksp2.north east, anchor=north east, fill=yellow!80, xshift=-3.2pt, minimum height=(\y1 - \y2 - 12pt - 0.2pt - 3.2pt)] { \scriptsize \color{gray} Node \\[-2pt] \color{gray} D };

\node [rosseq, right=0pt of vm.north east, anchor=center] { 3 }; %
\node [rosseq, right=0pt of ros dep.north east, anchor=center] { 4 }; %
\node [rosseq, right=0pt of vpn.south east, anchor=center] { 5 }; %
\node [rosseq, right=0pt of vpn.south east, anchor=center, xshift=23pt] { 5 };
\node [rosseq, right=0pt of wksp.north east, anchor=center] { 6 }; %
\node [rosseq, right=0pt of dds.south east, anchor=center] { 7 }; %
\node [rosseq, right=0pt of dds2.south east, anchor=center] { 7 }; %
\node [rosseq, right=0pt of node1.north east, anchor=center] { \tiny 11 };
\node [rosseq, right=0pt of node2.north east, anchor=center] { \tiny 11 };
\node [rosseq, right=0pt of robot node1.north west, xshift=4pt, anchor=center] { \tiny 12 };
\node [rosseq, right=0pt of robot node2.north west, xshift=4pt, anchor=center] { \tiny 12 };

\begin{pgfonlayer}{background}

\end{pgfonlayer}

\end{tikzpicture}
    \caption{\textbf{\FogROStwo{} Launch Sequence} This high-level overview shows the steps \FogROStwo{} takes in Section IV. Here we visualize a subset of the steps, keeping the same numbers as Section IV. The steps shown here are: (3) provision an instance, (4) install ROS and dependencies, (5) setup VPN, (6) copy ROS workspace to the instance, (7) setup DDS, (11) launch cloud-based nodes, and (12) launch robot nodes.  The gray nodes on the Robot are copied to the cloud computer and only launched in the cloud.  %
    }
    \label{fig:launch_sequence}
\end{figure}
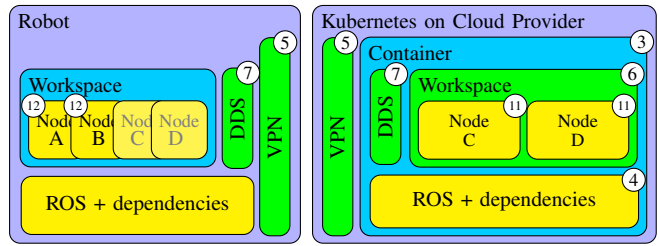

At the front end of \FogROStwo{} is the launch system that specifies what nodes to launch and \emph{where}.  Unlike FogROS1, \FogROStwo{}'s launch system is scriptable---allowing for launch-time logic to automate parts of the launch process.
Listing.~\ref{lst:launch} shows an example in which a grasp planner needs a GPU to run efficiently.  The script first defines a cloud machine with a GPU, then adds an attribute to the grasp planner node to tell the \FogROStwo{} launch process to run it on the cloud machine. In more extensive use cases, multiple nodes can run on the same machine, and \FogROStwo{} can launch multiple machines. %

The steps \FogROStwo{} takes are (\textbf{bold} items are new to \FogROStwo{}):
\begin{enumerate*}[label=(\arabic*)]
\item \textbf{trigger launch from integration with the command-line interface};
\item \textbf{process the launch script logic (e.g., to automate cloud selections, see Sec.~\ref{sec:launch_script_extensions})};
\item connect to the cloud provider through its programmatic interface to create and start a new instance along with setting up access control to isolate from other cloud computers, and generating secure communication key pairs, \textbf{using the new Kubernetes backend when needed};
\item install the ROS libraries and dependencies on the cloud machine needed for the robot application to run in the cloud (\textbf{skipped if using a pre-built custom image}, see Sec.~\ref{sec:image-management});
\item set up Wireguard virtual private networking (VPN) on robot and cloud machine to secure the \textbf{ROS\,2 DDS communication} between them;
\item copy the ROS nodes from the robot to the cloud machine;
\item \textbf{configure the DDS vendor's discovery mechanism to peer cloud and robot across the VPN};
\item \textbf{optionally configure streaming video compression (Sec.~\ref{sec:compression})};
\item launch docker instances;
\item \textbf{optionally launch Foxglove for monitoring (Sec.~\ref{sec:foxglove})};
\item launch cloud-based nodes;
\item launch nodes on the robot;
\item \textbf{once launched, users can use \FogROStwo{} CLI integrations interact with FogROS-related cloud computers (Sec.~\ref{sec:cli})}.
\end{enumerate*}

Once the launch process is complete, the nodes running on the robot and on the cloud machine(s) securely communicate and interact with each other---and the only change needed was a few lines of the launch script.  %

\subsection{ROS 2 Command Line Integration}
\label{sec:cli}

\FogROStwo{} integrates with the ROS\,2 Command Line Interface (CLI), offering an intuitive way to interact with \FogROStwo{} cloud instances not available in FogROS1 or \ROS1.  %
To use the CLI, a user types into a terminal window:
\[
\texttt{\footnotesize ros2 fog <command> [args...]}
\]
where \texttt{command} specifies an interaction with \FogROStwo{} along with additional arguments.  For example, \texttt{command} can be \texttt{list}, which lists cloud instances, \texttt{delete}, to delete existing instances that are no longer in use, \texttt{image}, to create and manage cloud images, or \texttt{connect}, to connect via SSH~\cite{rfc4251} to running instances.  

\subsection{Launch Script Extensions}
\label{sec:launch_script_extensions}

The new launch system in \FogROStwo{} enables custom logic during launch%
, which was 
not possible with the launch system in FogROS1.  We include several options:
\begin{itemize}
    \item Since the distance between robot and cloud can dramatically affect network latency, the launch script can select nearest cloud computer based on the robot location.  
    \item Since different computers (e.g., Intel vs Arm and with or without GPU) and regions require different images.  The launch script can automate selecting the correct image.
    \item Selecting the best or most cost-effective cloud computer for a \ROS2 node can require significant effort~\cite{yadwadkar2017selecting}.  By integrating ideas from Sky Computing~\cite{skycomputing}, the launch script can select a machine type based on a specification of requirements (e.g., CPU type and core count, memory size, GPU type and memory, and more).
\end{itemize}
As an example, changing Listing~1 line 5 to call \texttt{region= find\_nearest\_aws\_region()}, automates region selection.
This function uses the robot's IP and a geolocation API to determine the nearest cloud data center.
FogROS2 also offers a function to automate finding the cheapest instance type that matches a user's computing specifications.

\subsection{Cloud Computer Virtual Machine Image Management}
\label{sec:image-management}
Cloud computers with FogROS1 suffered from long startup times, sometimes approaching 4 minutes.  A significant portion of this time is due to installing ROS and dependencies on the cloud computer.  Advanced users could address this problem by creating computer images with software pre-installed.  \FogROStwo{} adds a tool to automate this process using the command-line interface, allowing it to start up pre-installed instances in a fraction of the time of FogROS1.

\subsection{New Kubernetes Backend}

FogROS1 supported AWS only.  To support additional cloud service providers,
\FogROStwo{} integrates a new %
Kubernetes~\cite{kubernetes} backend. %
Kubernetes (see Fig.~\ref{fig:vpc_overview_2}) is a system that orchestrates running \emph{containers}, %
or
units of software packages and their dependencies---e.g., a robot ROS node, \FogROStwo{}, ROS 2, and underlying operating system components. 
With Kubernetes, computers can already be on and waiting to run a new container.  This allows for significant speedup in startup time. There is a trade-off---machines managed by Kubernetes must already be on, and there can be significant initial delay when starting Kubernetes the first time.

\subsection{Streaming Image Compression}
\label{sec:compression}

\begin{figure}
    \centering
    \begin{tikzpicture}[node distance=6pt,>=stealth',font=\footnotesize,
   rosblock/.style={draw, rectangle, rounded corners, text width=38pt, align=center, inner sep=0, inner ysep=4pt}]
   
\node [inner sep=0pt] (robot label) { Robot };
\node [inner sep=0pt, right=3.4in of robot label.north west, anchor=north east, xshift=-7.5pt] (cloud label) {Cloud Instance};
   
\node [rosblock, fill=yellow, text width=40pt, minimum height=32pt, below=2pt of robot label.south west, anchor=north west] (camera node) {Camera \\ Node};
\node [rosblock, fill=yellow, text width=40pt, minimum height=32pt, right=3.4in of camera node.north west, anchor=north east, xshift=-7.5pt] (cv node) {Image Processor \\ Node};

\path [thick, ->, color=black] (camera node) edge node [align=center] (img topic) { \scriptsize \texttt{/camera} \\ (image topic)} (cv node);

\node [inner sep=0pt, below=26pt of camera node.south west, anchor=north west] (robot label 2) {Robot};
\node [rosblock, fill=yellow, text width=40pt, minimum height=42pt, below=2pt of robot label 2.south west, anchor=north west] (camera node 2) { Camera \\ Node };

\node [inner sep=0pt, right=3.4in of robot label 2.north west, anchor=north east, xshift=-7.5pt] (cloud label 2) { Cloud Instance };
\node [rosblock, fill=yellow, text width=40pt, minimum height=42pt, right=3.4in of camera node 2.north west, anchor=north east, xshift=-7.5pt] (cv node 2) {Image Processor \\ Node };

\node [rosblock, text width=42pt, right=6pt of camera node 2.south east, anchor=south west, fill=yellow, minimum height=42pt] (h264 pub) {};
\node [inner sep=0, below=3pt of h264 pub.north, anchor=north, text width=42pt, align=center] {\scriptsize H.264 Enc \\[-2pt] Node};
\node [rosblock, text width=32pt, above=4pt of h264 pub.south, anchor=south, fill=green] { \scriptsize H.264 \\[-2pt] encoder};

\node [rosblock, text width=42pt, left=6pt of cv node 2.south west, anchor=south east, fill=yellow, minimum height=42pt] (h264 sub) {};
\node [inner sep=0, below=3pt of h264 sub.north, anchor=north, text width=42pt, align=center] {\scriptsize H.264 Dec \\[-2pt] Node };
\node [rosblock, text width=32pt, above=4pt of h264 sub.south, anchor=south, fill=green] { \scriptsize H.264 \\[-2pt] decoder };

\path [thick, ->, color=black] (camera node 2) edge coordinate (c2e) (h264 pub);
\path [thick, ->, color=blue] (h264 pub) edge node [align=center, color=black] (vid topic) { \scriptsize \texttt{/camera/h264} \\ (video topic) } (h264 sub);
\path [thick, ->, color=black] (h264 sub) edge coordinate (d2c) (cv node 2);

\node [inner sep=0, below=28pt of c2e, align=center] (camimg1) { \scriptsize \texttt{/camera/src} \\ (remapped) };
\path [color=black, densely dotted] (camimg1) edge (c2e);

\node [inner sep=0, below=28pt of d2c, align=center] (camimg2) { \scriptsize \texttt{/camera} \\ (image topic) };
\path [color=black, densely dotted] (camimg2) edge (d2c);

\node [below=13pt of img topic, inner sep=1pt] (subcaption a) {\scriptsize (a) Before Image Transport Plugin };
\node [below=32pt of vid topic, inner sep=0pt] (subcaption b){\scriptsize (b) After Image Transport Plugin };

\begin{pgfonlayer}{background}

  \node [fit=(cloud label)(cv node)(cv node -| h264 sub.west),inner sep=0,draw=none,fill=none,xshift=-2pt,inner sep=4pt] (vpc) {};
  \node [fit=(cloud label 2)(h264 sub),inner sep=0,draw=none,fill=none,xshift=-2pt,inner sep=4pt] (vpc2) {};

  \path [draw]
      ([shift=(210:8pt)]vpc.210)
      to [bend left=80]
      ([shift=(160:8pt)]vpc.160)
      to [bend left=80]
      ([shift=(140:12pt)]vpc.140)
      to [bend left=60]
      ([shift=(110:6pt)]vpc.110)
      to [bend left=60]
      ([shift=(70:8pt)]vpc.70)
      to [bend left=60]
      ([shift=(50:8pt)]vpc.50)
      to [bend left=60]
      ([shift=(0:8pt)]vpc.0)
      to [bend left=60]
      ([shift=(320:8pt)]vpc.320)
      to [bend left=60]
      ([shift=(290:6pt)]vpc.290)
      to [bend left=60]
      ([shift=(260:8pt)]vpc.260)
      to [bend left=60]
      ([shift=(240:8pt)]vpc.240)
      to [bend left=60]
      cycle;
      
    \path [draw, fill=white]
      ([shift=(210:8pt)]vpc2.210)
      to [bend left=80]
      ([shift=(160:8pt)]vpc2.160)
      to [bend left=80]
      ([shift=(140:12pt)]vpc2.140)
      to [bend left=60]
      ([shift=(110:6pt)]vpc2.110)
      to [bend left=60]
      ([shift=(70:8pt)]vpc2.70)
      to [bend left=60]
      ([shift=(50:8pt)]vpc2.50)
      to [bend left=60]
      ([shift=(0:8pt)]vpc2.0)
      to [bend left=60]
      ([shift=(320:8pt)]vpc2.320)
      to [bend left=60]
      ([shift=(290:6pt)]vpc2.290)
      to [bend left=60]
      ([shift=(260:8pt)]vpc2.260)
      to [bend left=60]
      ([shift=(240:8pt)]vpc2.240)
      to [bend left=60]
      cycle;
      
   \node [fit=(subcaption a), fill=white, inner sep=0pt, rounded corners] {};
      
   \node [fit=(robot label 2)(h264 pub),draw=black,rounded corners,fill=blue!30] {};
   \node [fit=(cloud label 2)(h264 sub),draw=black,rounded corners,fill=blue!30] (cloud box 2){};

   \node [fit=(robot label)(camera node)(camera node -| h264 pub.east),draw=black,rounded corners,fill=blue!30] {};
   \node [fit=(cloud label)(cv node)(cv node -| h264 sub.west),draw=black,rounded corners,fill=blue!30] (cloud box) {};

\end{pgfonlayer}

\end{tikzpicture}
    \caption{\textbf{\FogROStwo{} Streaming Video Compression} Unlike FogROS1, \FogROStwo{} uses streaming video compression of image topics to reduce bandwidth and latency of cloud-based image processing.  In (a), a cloud-based image processor node subscribes to the camera without streaming compression.  In (b), \FogROStwo{}, using the Image Transport Plugin, introduces an H.264 encoder (compression) node on the robot, and pairs it with an H.264 decoder (decompression) node on the cloud.  In both, the cloud-based Image Processor Node subscribes to the same topic. %
    }
    \label{fig:image_transport}
\end{figure}
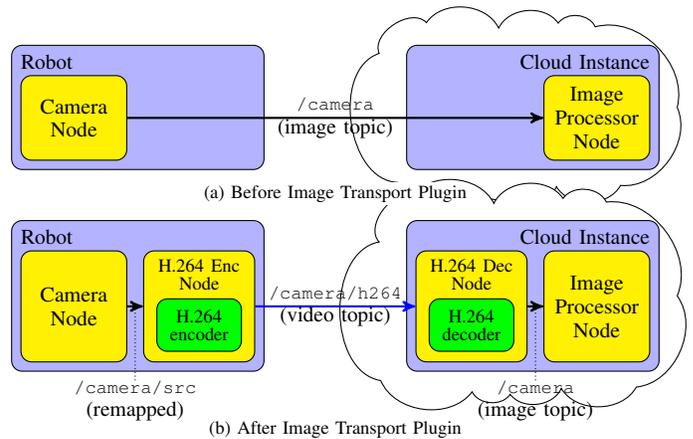

Many robot algorithms depend on fast processing of image and video data, and these algorithms increasingly require hardware acceleration e.g., via GPUs.  However, images and videos are data intensive in ROS, and the time to transmit data to the cloud can reduce the advantage of cloud-based acceleration.  Processing images in the cloud was possible with FogROS1, but with high latency.

To address this, at launch time \FogROStwo{} can setup transparent streaming compression between robot and cloud. The video compression we use is H.264~\cite{richardson2011h} from the open-source libx264~\cite{libx264} library.  Using H.264 allows \FogROStwo{} to greatly reduce the latency of processing video in the cloud.  We implement a ROS\,2 \emph{image\_transport\_plugin}~\cite{imagetransportplugins} to make the compression transparent to the application---publishing nodes still publish a sequence of images and subscribing nodes still receive a sequence of images.  %
Fig.~\ref{fig:image_transport}~(b) shows how \FogROStwo{} implements transparent streaming compression.

H.264 compression is %
often hardware accelerated, reducing the CPU utilization required to compress and decompress.

\subsection{Remote Monitoring and Visualization with Foxglove}
\label{sec:foxglove}

Users of robots running FogROS1 were only able to monitor robots locally, missing out on an advantage of the cloud.
\FogROStwo{} integrates Foxglove~\cite{foxglove}---a browser-based tool that enables visualization of ROS 2 topics. 
Much like rviz, the 3D visualizer that is part of ROS, Foxglove operates by subscribing to ROS messages, then interpreting and displaying them.  The chief difference is that Foxglove runs in a browser, and thus requires messages to be transmitted over web-based protocols.  \FogROStwo{} uses two components to integrate Foxglove: (1) a Foxglove server, that provides the web interface software, and (2) \emph{ROS bridge}, a ROS\,2 node that subscribes to topics as a ROS node and proxies them through websockets to a browser running the Foxglove software (Fig.~\ref{fig:vpc_overview_2} bottom).  When visualizations are enabled, \FogROStwo{} launches both the Foxglove server as a docker, and the ROS bridge node.

Once set up, \FogROStwo{} provides the IP address of the server, allowing multiple users in different locations to monitor and visualize the robot application in a browser. %

\section{Evaluation}
\label{sec:evaluation}

\begin{figure}
    \centering
    \includegraphics[width=\linewidth]{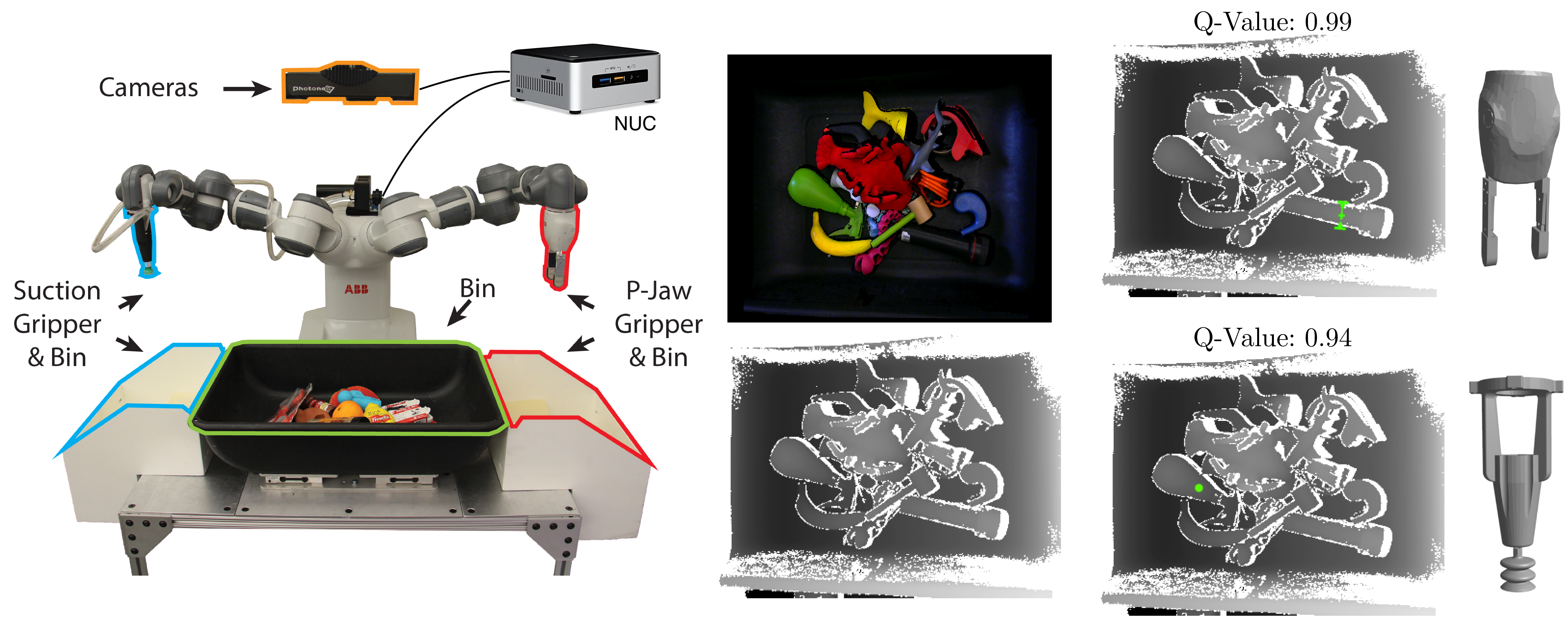}
    \caption{\textbf{Grasp Planning with Dex-Net}
    A robot (left) with an attached computer (e.g., an Intel NUC) sends RGBD image observations from an overhead camera (middle) to compute grasps using a Dex-Net Grasp Quality Convolutional Neural Network (GQCNN) on a cloud computer with a GPU.  The result (right) is grasps poses (green) and their relative quality score (q-value) for parallel-jaw (top-right) and suction gripper (bottom-right). %
    }
    \label{fig:dex-net_example}
\end{figure}

We evaluate the ability of \FogROStwo{} to speed up robot computations using the cloud, to lower startup latency, to lower network latency and utilization.

We use an Intel NUC with an Intel\textsuperscript\textregistered{} Pentium\textsuperscript\textregistered{} Silver J5005 CPU @ 1.50\,GHz with 2 cores enabled and with a 10\,Mbps network connection to act as the robot, as it is representative of an efficient computing platform that could be onboard or attached to a robot (Fig.~\ref{fig:dex-net_example}).  We perform all evaluations with cloud nodes deployed to AWS unless specified otherwise.  This setup differs from the examples in FogROS1, thus we re-run the experiments in FogROS1~\cite{chen2021fogros} to compare FogROS1 and FogROS2 on equivalent hardware. 

\subsection{Streaming Video Compression}

\begin{table}[t]
    \centering
    \footnotesize
    \begin{tabular}{@{}l@{\quad}r@{\quad}r@{\quad}r@{}}\toprule
         Mode & FPS & Latency (ms) \\
         \midrule
         Uncompressed  & 1.42 & 1401 \\
         Compressed    & 15.0 & 333 \\
         Theora        & 29.6 & 83 \\
         H.264         & 29.0 & \textbf{38} \\
         \bottomrule
    \end{tabular}
    \caption{\textbf{Streaming \FogROStwo{} video compression} A node on a robot publishes 3000 images to a ROS topic that \FogROStwo{} transparently compresses and sends to a node in the cloud.  We measure the frames per second (FPS) the cloud node receives. %
    For every image, the cloud node publishes a small acknowledgement message that the robot uses to measure the round-trip time, or latency (ms).}
    \label{tab:streaming_video}
\end{table}

We evaluate the performance of using streaming H.264 video compression between robot and cloud.  In this experiment, we have the robot node publish images to a topic to which a cloud node subscribes.  The cloud node responds immediately with a small acknowledgement message.
We record the round-trip time, %
and frames per second (FPS), and show it in Table~\ref{tab:streaming_video}. 
We compare to \textbf{Uncompressed}, which is the raw pixel arrays native to \ROS1 and \ROS2, \textbf{Compressed}, which uses (non-streaming) image compression, \textbf{Theora} ~\cite{theora2017} streaming image compression, and \FogROStwo{}'s H.264 compression.
From the table, we observe the benefit of streaming video compression between the robot and the cloud, as the cloud can receive images 13$\times$ faster FPS, while %
shortening the latency by 97\,\%.  We observe performance improvement of H.264 over Theora, the previous best available compression, with H.264 %
shortening the latency by 54\,\%.  The reduced latency from 1401\,ms to 38\,ms may enable some real-time cloud-robotics applications not possible without compression.

However, there may be a tradeoff in some applications.  Theora and H.264 are both lossy compression algorithms, meaning they are designed to compress videos by discarding some image information.  This information loss is tailored to human perception~\cite{richardson2011h}, and thus may adversely affect computer vision algorithms.

\subsection{Cloud Robotics Benchmark Applications}

We evaluate \FogROStwo{} in a benchmark on 3 example robot applications: SLAM with ORB-SLAM2~\cite{mur2017orb}, Grasp Planning with Dex-Net~\cite{mahler2017dex}, and Motion Planning with Motion Planning Templates (MPT)~\cite{ichnowski2019mpt}%
.  Refer to FogROS1~\cite{chen2021fogros} for further details on these benchmarks.
We compare to a baseline of robot-only computing and FogROS1 using equivalent cloud computers.
For examples with cloud-based image processing, we compare to additional baselines of (a) raw/uncompressed, (b) PNG compressed, and (c) Theora~\cite{theora2017} compressed, where Theora is an open-source video compression library with an existing image transport plugin~\cite{imagetransportplugins}. 

\begin{table}[t]
    \centering
    \footnotesize
    \begin{tabular}{@{}lccc@{\;}c@{}}\toprule
                  & Robot & FogROS1 
                  & \multicolumn{2}{c}{ \FogROStwo{} }
                  \\
                  \cmidrule{4-5} 
         Scenario & Only & Compressed & Compressed & H.264 \\
         \midrule
         fr1/xyz & 0.52 & 1.62 & 0.82 & \bf 0.24 \\
         fr2/xyz & 0.43 & 1.61 & 0.75 & \bf 0.25 \\
         fr2/loop & 0.68 & 1.63 & 0.89 & \bf 0.22 \\
         \bottomrule
    \end{tabular}
    \caption{\textbf{ORB-SLAM2 results on \FogROStwo{}}
    We run ORB-SLAM2~\cite{mur2017orb} on 3 benchmarks from the TUM Dataset~\cite{sturm12iros} and measure average per-frame round-trip latencies incurred by the ORB-SLAM2 node. Here, \FogROStwo{} runs on a 36-core cloud computer.  H.264 compression allows \FogROStwo{} to outperform robot-only and image-compressed (not video-compressed).}
    \label{tab:slam}
\end{table}
\begin{table}[t]
    \centering
    \footnotesize
    \begin{tabular}{@{}l@{\quad}c@{\quad}c@{\quad}r@{\quad}r@{\quad}r@{\quad}r@{}}\toprule
                  & Robot & Cloud &
                  \multicolumn{2}{c}{FogROS1} &
                  \multicolumn{2}{c}{\FogROStwo{}} \\
                  \cmidrule(r){4-5}
                  \cmidrule{6-7} 
         Scenario & Only & Compute & Network & Total & Network & Total \\
         \midrule
         Uncompressed & 14.0 & 0.6 & 5.0 & 5.6 & 5.0 & 5.6 \\
         Compressed   & 14.0 & 0.6 & 1.3 & 1.9 & 0.7 & 1.3 \\
         H.264        & 14.0 & 0.6 & -   &   - & 0.6 & \bf 1.2 \\
         \bottomrule
    \end{tabular}
    \caption{
    \textbf{Dex-Net results on \FogROStwo{}}
    We measure compute time in seconds for 10 trials on a robot with a CPU, and compute and network time using cloud computer with an Nvidia T4 GPU via \FogROStwo{}.}
    \label{tab:dexnet}
\end{table}

\begin{table}[t]
    \centering
    \footnotesize
    \begin{tabular}{@{}l@{\quad}c@{\quad}c@{\quad}r@{\;}r@{\quad}r@{\;}r@{\;}r@{}}\toprule
                  & Robot & Cloud & 
                  \multicolumn{2}{c}{FogROS1} & 
                  \multicolumn{3}{c}{\FogROStwo{}} \\
                  \cmidrule(r){4-5} \cmidrule{6-8}
         Scenario   & Only &  Compute & Network & Total & Network & Total & Speedup \\
         \midrule
         Apartment  & 157.6 & 3.46 & 0.07 & 3.53 & 0.04 & 3.50 & 45$\times$ \\
         Cubicles   & \phantom035.8 & 1.51 & 0.07 & 1.58 & 0.05 & 1.56 & 23$\times$ \\
         Home       & 161.8 & 4.73 & 0.08 & 4.81 & 0.05 & 4.78 & 34$\times$ \\
         TwistyCool & 167.9 & 4.76 & 0.08 & 4.84 & 0.05 & 4.81 & 35$\times$ \\
         \bottomrule
    \end{tabular}
    \caption{
    \textbf{MPT Motion Planning results on \FogROStwo{}}  We run multi-core motion planners from Motion Planning Templates (MPT)~\cite{ichnowski2019mpt} on 4 motion planning problems from the Open Motion Planning Library OMPL~\cite{OMPL}.%
    We record the planning time running on a 96-core cloud computer, and the network round-trip time between robot and cloud.}
    \label{tab:mpt}
\end{table}

In SLAM experiment (Table~\ref{tab:slam}), H.264 compression lowers per-frame latency from 0.82\,s to 0.24\,s, a 3.4$\times$ improvement, allowing FogROS2 to achieve lower latency than robot-only computation.  In Dex-Net experiments (Table~\ref{tab:dexnet}), FogROS2 and H.264 results in a $12\times$ speedup.  In MPT experiments (Table~\ref{tab:mpt}), planning speeds up by up to $45\times$ (FogROS1 also speeds up similarly due to hardware changes in experiments).

\subsection{Cloud-Computer Startup Times}

We test if \FogROStwo{} can shorten startup times compared to FogROS1.  Short startup times benefit software development cycles and robots that intermittently operate (e.g., vacuuming robots).  In this experiment, we use the new $\texttt{image}$ command in \FogROStwo{} to generate a custom computer image, and measure the time between launch and first robot-cloud \ROS2 node interaction with and without the custom image.  For comparison, we manually create a custom image for FogROS1 by using the AWS web console.

\begin{table}[t]
    \centering
    \begin{tabular}{@{}lr@{\quad}r@{\quad}r@{\quad}r@{}}\toprule
                        &         & \multicolumn{3}{c}{\FogROStwo{}} \\
                        \cmidrule(lr){3-5}
         Computer Image & FogROS1 & AWS & GCP (K8s) & Local (K8s) \\
         \midrule
         Default & 228$\pm$25\,s\phantom{*}  &  275$\pm$61\,s & \multirow{2}{*}{\textbf{29$\pm$2.1\,s}$^\text{\bf b}$} & \multirow{2}{*}{\textbf{27$\pm$0.83\,s}$^\text{\bf b}$} \\
         Custom &  155$\pm$32\,s$^\text{\bf a}$  &  \textbf{85$\pm$11\,s} \\
         \bottomrule
         \multicolumn{3}{l}{\vphantom{$X^X$}\footnotesize $^\text{\bf a}$\,FogROS1 custom image manually created} \\
         \multicolumn{3}{l}{
         \footnotesize $^\text{\bf b}$\,Kubernetes cluster already started.
         }
    \end{tabular}
    \caption{\textbf{Startup time for cloud computer}  \FogROStwo{} automates the creation of custom computer images with pre-installed ROS and dependencies.  This speeds up cloud-computer startup allowing the robot to operating sooner.  Kubernetes uses containers that have many of the ROS component already installed---thus they save time installing ROS, but are not fully customized like the AWS images.}
    \label{tab:startup_times}
\end{table}

\begin{table}[t!]
    \centering
    \begin{tabular}{@{}lrr@{}}\toprule
         & west coast & east coast \\
         \midrule
    us-west-1 & \textbf{6.1\,ms} & 72\,ms \\
    us-east-2 & 74\,ms & \textbf{13\,ms} \\
    \bottomrule
    \end{tabular}
    \caption{Example round-trip data times based on robot location (west or east coast) and the cloud computer's data center (us-west-1 vs us-east-2).  \FogROStwo{} launch extension can automatically select the nearest data center resulting in lower network latency.}
    \label{tab:region_selection}
\end{table}

In Table~\ref{tab:startup_times}, we observe that the custom image in \FogROStwo{} reduces AWS startup times by 63\,\%.

Using Kubernetes on local cluster or Google Cloud Platform (GCP) reduces startup times further, due to having the computers in the Kubernetes cluster already running.  The AWS backend must create a new cloud computer each time, accounting for approximately 40\,s of delay. 

There is a tradeoff to be made in startup times.  Kubernetes requires starting up a cluster of computers, which can take on the order of 10 minutes.  If one is willing to spend this time upfront, Kubernetes allows users to redeploy ROS nodes over and over, which may be beneficial when rapidly prototyping changes to robot code.

\subsection{Automating Region Selection}

We test the launch script extensions for automatic region selection to allow robots to select the cloud data center that is nearest to them.  We test deploy a robot on the US west coast, and the launch script selects the \texttt{us-west-1} data center.  When we test to deploy the robot on the US east cost, the robot selects the \texttt{us-east-2} data center. Table~\ref{tab:region_selection} shows example round-trip network times in this experiment, with the best latencies in bold and automatically selected by the geolocation script.

\section{Conclusion}
\label{sec:conclusion}

We present \FogROStwo{}, an adaptive cloud-robotics platform for running compute-intensive portions of ROS\,2 applications in the cloud.  \FogROStwo{} addresses 9 shortcomings of FogROS1, integrates with the ROS\,2 launch and communication systems, to provision and start cloud computers, configure and secure network communication, install robot code and dependencies, and launch robot and cloud-robotics code.  As a redesigned and distinct successor to FogROS, \FogROStwo{} supports ROS 2, transparent video compression, improved performance and security, access to more cloud computing providers, and remote visualization and monitoring.  In experiments, we observe a significant performance benefit to using cloud computing, with the additional improvement from transparent video compression. %

In future work, we will continue to improve performance and capablities of FogROS2.  %
We will explore additional models of computing, such as serverless, spot instances, and more.  We will also explore extending the networking capabilities of FogROS2 to allow multiple robots to communicate, collaborate, and share data more easily~\cite{chen2022fogrosg}.

\section*{Acknowledgements}

This research was performed at the AUTOLab at UC Berkeley in
affiliation with the Berkeley AI Research (BAIR) Lab, the CITRIS ``People and Robots'' (CPAR) Initiative, the Real-Time Intelligent Secure Execution (RISE) Lab, and by the NSF/VMware Partnership on Edge
Computing Data Infrastructure (ECDI) Secure Fog Robotics Project Award 1838833.
We thank our colleagues for helpful discussions and testing.
We thank Brandon Fan and Emerson Dove for their contributions and discussion related to the new Kubernetes backend.

\bibliographystyle{IEEEtran}
\bibliography{IEEEabrv,references}

\end{document}